\documentclass[11pt,a4paper]{article}
\usepackage[hyperref]{naaclhlt2018}
\usepackage{times}
\usepackage{latexsym}
\usepackage{dirtytalk}

\usepackage{url}

\aclfinalcopy 

\title{Absit invidia verbo: Comparing Deep Learning methods for offensive language}

\author{Bogdan Lazarescu \\
    Imperial College London \\
   bogdanblazarescu@gmail.com\And
  Christo Lolov \\
  Imperial College London \\
   christo.lolov@gmail.com\And
  Silvia Sapora \\
  Imperial College London \\
  silvia.sapora@gmail.com}

\begin{document}
\maketitle
\begin{abstract}
 This document describes our approach to building an Offensive Language Classifier. More specifically, the OffensEval 2019\cite{offenseval} competition required us to build three classifiers with slightly different goals:
  \begin{enumerate}
    \item Offensive language identification: would classify a tweet as offensive or not.
    \item Automatic categorization of offense types: would recognize if the target of the offense was an individual or not.
    \item Offense target identification: would identify the target of the offense between an individual, group or other.
  \end{enumerate}
  In this report, we will discuss the different architectures, algorithms and pre-processing strategies we tried, together with a detailed description of the designs of our final classifiers and the reasons we choose them over others.
  We evaluated our classifiers on the official test set provided for the OffenseEval 2019 competition\cite{OLID}, obtaining a macro-averaged F1-score of 0.7189 for Task A, 0.6708 on Task B and 0.5442 on Task C. 
\end{abstract}

\section{Pre-Processing}
Our pre-processing included:
  \begin{enumerate}
    \item Removing all \say{@USER} strings
    \item Removing all \say{URL} strings
    \item Removing all punctuation
    \item Removing all symbols
    \item Converting all text lowercase
    \item Expanding abbreviations
  \end{enumerate}
 Other steps we considered were:
    \begin{enumerate}
    \item Removing all emojis
    \item Removing all hashtags
    \item Removing all numbers
    \item Removing stop words
  \end{enumerate}
 After testing, we determined those changes either did not contribute positively or decreased the performance of our classifiers. The resulting corpus is then tokenized using the Natural Language Toolkit (nltk\cite{Loper02nltk:the}) library, in order to obtain the individual composing words, that will later be embedded and used as inputs for the classification models.
\section{Input Data}
Initially, we decided to use the Bag-of-Words model. To do this, we used \texttt{sklearn}'s\cite{scikit-learn} function \texttt{CountVectorizer}. \texttt{CountVectorizer} converts a series of text strings into a matrix of token counts.
This implementation produces a sparse representation of the counts using \texttt{scipy.sparse.csr\_matrix}. We had the option to either provide \texttt{CountVectorizer} with an a-priori dictionary or to let the library create one using the words provided in the given input data.\par
As word embeddings have the ability to generalize (thanks to semantically similar words having similar vectors) \cite{medium-word-embeddings} we decided we would try out this approach and check its performance. We used the \texttt{Word2Vec} function from the \texttt{gensim}\cite{gensim} library to get word embeddings we could then use to train a model with. Our approach was to first get a dictionary mapping each word to a n-dimensional vector (we tried n=100 to start with). Then, to build the feature vector, we tried to average word vectors for all the words in a given sentence. To do this we built a sklearn-compatible transformer that is initialized with a word to vector dictionary. Unfortunately, we had some problems getting the results from the \texttt{Word2Vec} function and averaging them. \texttt{gensim} provides many methods to calculate the similarity of words and other properties, but it doesn't provide any direct way of getting a list of features for each sentence. This is why we decided to abandon this approach. \par 
We also tried using Term Frequency-Inverse Document Frequency (TF-IDF) through the \texttt{sklearn} function \texttt{TfidfVectorizer}. Surprisingly, it yielded worse results than Bag of Words when we tested it with our Logistic Regression classifier. \par 
For our Neural Networks(RNNs and CNN) we decided to use the \texttt{word2idx} function we implemented. This function would first assign each word an index, an then it would replace each word in a sentence with its corresponding index. This meant our sentences were now list of integers.

\section{Performance and Evaluation}
In order to train and evaluate the performance of our classifiers, we used the provided training dataset together with the validation one. During training, we split the 13240 tweets in a 90-10 ratio where the 10\% was used for validation in between epochs. The provided validation dataset was in fact used for testing our model. For each model, we compared the predictions on the test data to the provided reference classifications baseline, using accuracy and F1 with macro averaging as metrics.

\section{Libraries and Frameworks}
In our implementations we made extensive use of PyTorch \cite{paszke2017automatic}, Keras \cite{chollet2015keras}, scikit-learn, and Natural Language Toolkit.\par
We decided to use Natural Language Toolkit because it had a base Tokenizer for Tweets. The Tokenizer was a good starting point because it took care of removing handles, allowed easy lower casing and stripped repeating symbols to a certain length. As the coursework progressed, we built on top of it while experimenting with pre-processing of sentences. 
\par
When we considered deep learning frameworks we looked into TensorFlow \cite{tensorflow2015-whitepaper}, Keras and PyTorch. Both TensorFlow and Keras are older frameworks than PyTorch. Keras has a reputation for being easy to reason about, while TensorFlow has the reputation of more powerful, used in production systems and having a steep learning curve. PyTorch, on the other hand, seemed to combine the best of both by being easy to reason about, unlike TensorFlow, and yet enabling developers to go into low-level implementation details, unlike Keras. We chose to go with PyTorch when implementing the CNN because the majority of our team had already used it for other courses and we felt confident working with it. We mainly used Keras for the RNNs to compare whether it was indeed easier to use than PyTorch.\par
We chose to use the linear regression model in scikit-learn over that in PyTorch because we thought the API was more intuitive to use.

\subsection{Convolutional Neural Network (CNN)}
As our first attempt, we decided to approach the classification problem by using a Convolutional Neural Network.\par
The CNN which gave us the best results used an embedding, a feature and a pooling layer. It made use of a \texttt{word2idx} encoding of tweets padded to the length of the longest tweet. The CNN favored a 100 by 1 vector for its embedding representation which was then sampled using a window of size 3 in order to extract useful features. The loss function which gave best results was \texttt{BCEWithLogitsLoss} and the optimizer which worked in conjunction with it was \texttt{Adam}.\par
In the beginning, we began with a window size of 1 i.e. no neighbourhood around the word was used to enhance its semantics. By expanding the window size and plotting it against the accuracy and F1 scores we saw that they leveled off at a window size of 3.\par
As a next step, we tried to see whether increasing size of the output vector from the embedding layer would improve the feature extraction. We went up to 1000 entries in a one-dimensional vector, but that drastically added to the computational time and barely altered the output scores.\par
Afterward, we experimented with changing the structure of the network and making it deeper. We added a couple more convolutional layers, but the network still only reached an accuracy of 66\%.\par
We made a similar experiment by varying the probability of a feature being dropped   (originally set to 50\%). It turned out, however, that both increasing and decreasing the probability caused a deterioration in our results rather than an improvement.\par
At this point, we considered the structure of the dataset. For task A, the dataset was biased because 66\% of the data classified as NOT offensive, while 33\%  classified as OFFensive. This meant that the network could have learned to achieve the accuracy and F1 results we obtained just by classifying everything as the predominant class i.e. NOT offensive.\par
In order to mitigate this issue, we tried experimenting with different loss functions and optimizers. We started off using \texttt{BCEWithLogitsLoss}, which weights all classes as if each had an equal portion of the dataset. We tested other loss functions (\texttt{MSELoss}, \texttt{CrossEntropyLoss}) which we have seen to have a good performance in other deep learning scenarios, but they did not yield better results.\par
The optimizer which we originally used was \texttt{SGD}, but we found out that the optimizer which worked the best with \texttt{BCEWithLogitsLoss} was \texttt{Adam}.\par
After altering pretty much every parameter we could and not achieving a significant improvement in our results, we decided that a CNN might not be the best approach to this problem and considered using Recurrent Neural Networks.

\subsection{Recurrent Neural Network (RNN)}
Having in mind all the previous results, we decided to move our attention to RNNs \cite{pak2010twitter}. Considering the particularities of RNN models that make them suitable for working with sequential data, we decided to use their capabilities to solve our classification task.\par
We made use of two different approaches: LSTM and Bidirectional LSTM (B-LSTM) which try to mitigate some of the vanilla RNN drawbacks, mainly, the vanishing gradient \cite{tai2015improved}.\par
Since both LSTM and B-LSTM gave us results above the baseline we decided to used them for tasks B and C as well. We trained different models for each task. After the input was pre-processed and tokenized, we used the \texttt{word2idx} function we defined to get a sentence embedding. We then padded the result with 0 values to obtain equally sized inputs.\par The resulted embeddings were used for both the LSTM and the B-LSTM models which share a common architecture. The input layer of both of them is connected to an Embedding layer whose goal is to reduce the input's dimensionality into a more meaningfully latent space which facilitates the classification procedure. The next layer is composed of LSTM or B-LSTM cells. The two share a common internal structure, but the difference lies in the B-LSTM: it is built out of two LSTM cells, one is trained forward (from left to right, the natural order of the input sequence), while the other is trained backward (from right to left, considering the inverse order of the input). The output is a result of a Dense Layer which takes the result of the above LSTM/B-LSTM and reduces it to the probability of the input belonging to one of the output classes.

\subsection{Logistic Regression (LR)}
We were quite interested to see how Neural Networks perform against a simpler method like Logistic Regression. As such, we trained a model for each task in order to compare its behaviour to our RNNs.\par
We started off by creating a one-hot encoding of the training dataset \cite{agarwal2011sentiment}. Upon making predictions on the testing dataset we ran into the problem where the one-hot encoding of the testing dataset was different to the one-hot encoding of the training dataset. This was due to the fact that not all words in the training dataset were seen in the testing dataset. We worked around this problem by using the testing dataset in conjunction with the training dataset when creating the one-hot encoding. When we trained the Logistic Regression model we only fed it the one-hot encoding of the training dataset.\par
To our surprise, Logistic Regression appeared to perform quite well on all three tasks.

\section{Training Challenges and Tuning}
During the training stage, a series of optimization decisions were made. Firstly, as the corpus dimension was not very big and suffered from imbalance, we decided to use a network with just 4 layers (Input, Embedding, LSTM/B-LSTM, Dense=Output). This measure was taken in order to reduce the over-fitting and allowed us to use LSTM for the given task. We tried adding more layers, but unfortunately, the increased number of parameters inevitably lead to poor generalization capabilities, as the higher flexibility of the model adapted too well to the training data it was fed. The Embedding layer tries to compress the 2500 words (vocabulary size) into an embedding vector of size 60. We chose to limit ourselves to the first 2500 words from our vocabulary as we observed, after plotting the word frequency, that the occurrence frequency dropped dramatically after the first 2486 words. At the next stage, the LSTM/B-LSTM layer takes its input from the Embedding layer, a vector of 60 elements, and produces a cell output of size 100. After repeated experiments, we found 100 was the dimension that offered the best result for the given problem. This output goes to the Dense layer which predicts the probability of the input sentence being Offensive/NotOffensive, Targeted/NotTargeted and Individual/Collective/Other, depending on the problem task.\par
During the tuning of hyperparameters, we built different models with different properties. Plotting the training versus validation accuracy we observed that the models were highly over-fitting. As we couldn't enhance the training set, we tried other regularization techniques. The first one was Dropout, which showed its usefulness by reducing the over-fitting and increasing the accuracy on the validation set by about 1.5\%. Next, we tried changing the output size of the some of the network's layers (Embedding size and LSTM memory cell size) which in turn brought some improvement. Another technique we used with a noticeable improvement on the model's performance was L2 regularization: it slightly increased the generalization capabilities of the network and reduced the discrepancy between the training set accuracy and the test one. At last, we plotted the accuracy versus the number of epochs used for training and we used Early Stopping to maximize our generalization capabilities.\par
As the B-LSTM considers the sequence from both ends, it gets a better consideration of the context. These extra capabilities enhanced the performance, for our tasks, with an average of 3\%, compared to a similar LSTM model. During hyperparameter tuning, we observed that given the reduced size of the corpus the B-LSTM overfits easily, so the early stopping was used to mitigate those effects. 

\section{Results}
For task A, we tested many different models and compared their performance. From the results we got, we concluded Logistic Regression, LSTM and B-LSTM were the best ones. Because of this, we decided to only use them on the following tasks and to spend our time trying to improve their performance by trying out different hyperparameters. We also submitted the predictions for all the models we used to Codalab and waited for the results to make the final decision on which models worked the best. Table \ref{table:table1} shows the accuracy and F1 score for each model we tested, while Table \ref{table:table2} summarizes how those models performed on the official test set.

\begin{table}[]
\begin{tabular}{lll}
\hline
\multicolumn{3}{|l|}{\bf Task A}\\ \hline
\multicolumn{1}{l|}{Model} & 
\multicolumn{1}{l}{\bf Accuracy} & 
\multicolumn{1}{|l}{\bf F1} \\ \hline
B-LSTM 20 epochs & 0.7190 & 0.6901 \\
B-LSTM 5 epochs & 0.7462 & 0.7147 \\
B-LSTM 7 epochs + L2 & 0.7892 & 0.7309 \\
LSTM 6 epochs + L2 & 0.7522 & 0.7165 \\
LSTM 12 epochs + L2 & 0.7322 & 0.6322 \\
CNN & 0.6148 & 0.34\\
Multinomial NB + RD & 0.7726 & 0.7687\\
SGD Classifier + RD & 0.7794 & 0.7781\\
Logistic Regression & 0.7681 & 0.7282\\
Logistic Regression + RD & 0.8529 & 0.8528\\\hline
\multicolumn{3}{|l|}{\bf Task B}\\ \hline
\multicolumn{1}{l|}{Model} & \multicolumn{1}{l}{\bf Accuracy} & \multicolumn{1}{|l}{\bf F1} \\ \hline
LSTM & 0.6681 & 0.3965 \\
B-LSTM & 0.5681 & 0.4399 \\
Logistic Regression & 0.85 & 0.5247\\
Logistic Regression + RD & 0.9072 & 0.9067 \\\hline
\multicolumn{3}{|l|}{\bf Task C}\\ \hline
\multicolumn{1}{l|}{\bf Model} & 
\multicolumn{1}{l}{\bf Accuracy} & 
\multicolumn{1}{|l}{\bf F1} \\ \hline
LSTM & 0.6881 & 0.4929 \\
B-LSTM & 0.7233 & 0.5023 \\
Logistic Regression & 0.7164 & 0.4630 \\
Logistic Regression + RD & 0.8478 & 0.8456 \\\hline
\end{tabular}
\caption{Tasks performance of different models. Each model is trained with 90\% of training data and tested on the remaining 10\%. RD indicates Random Draw to balance the number of samples of each type. L2 indicates L2 regularization.}
\label{table:table1}
\end{table}

\begin{table}[]
\begin{tabular}{ll}
\hline
\multicolumn{2}{|l|}{\bf Task A}\\ \hline
\multicolumn{1}{l}{Model} & 
\multicolumn{1}{|l}{\bf F1} \\\hline
LSTM 8 epochs + L2 & 0.718944681 \\
B-LSTM 7 epochs + L2 & 0.72103681 \\
Logistic Regression + RD & 0.72552438 \\\hline
\multicolumn{2}{|l|}{\bf Task B}\\ \hline
\multicolumn{1}{l}{Model} &
\multicolumn{1}{|l}{\bf F1} \\ \hline
LSTM 7 epochs + L2 & 0.542312321 \\
B-LSTM 6 epochs + L2 & 0.560707749 \\
Logistic Regression + RD & 0.670806302 \\\hline
\multicolumn{2}{|l|}{\bf Task C}\\ \hline
\multicolumn{1}{l|}{\bf Model} & 
\multicolumn{1}{l}{\bf F1} \\ \hline
LSTM 7 epochs + L2 & 0.439543212 \\
B-LSTM 6 epochs + L2 & 0.441269841\\
Logistic Regression + RD & 0.544192563 \\\hline
\end{tabular}
\caption{Tasks performance of different models on the official test set.}
\label{table:table2}
\end{table}

\section{Conclusion}
This task provided a good hands-on experience with Natural Language Processing. Our main takeaway was that suitably trained simpler models (Logistic Regression) can sometimes perform as well, if not better, than more complicated ones (RNNs).\par
Our code and resources can be accessed via the following Google Drive URL: \url{https://drive.google.com/drive/folders/10DDRyFcQ2cszSZAwnP2Lu-Bf4TB89urP}

\bibliography{semeval2018}
\bibliographystyle{acl_natbib}

\appendix

\end{document}